\documentclass[10pt,twocolumn,letterpaper]{article}

\usepackage[pagenumbers]{cvpr} 

\usepackage{graphicx}
\usepackage{amsmath}
\usepackage{amssymb}
\usepackage{booktabs}

\usepackage[pagebackref,breaklinks,colorlinks]{hyperref}

\usepackage[capitalize]{cleveref}
\crefname{section}{Sec.}{Secs.}
\Crefname{section}{Section}{Sections}
\Crefname{table}{Table}{Tables}
\crefname{table}{Tab.}{Tabs.}

\def\cvprPaperID{*****} 
\def\confName{CVPR}
\def\confYear{2022}

\begin{document}

\title{Accelerating Production LLMs with Combined Token/Embedding Speculators}

\author{Davis Wertheimer, Joshua Rosenkranz, Thomas Parnell, Sahil Suneja,\\
Pavithra Ranganathan, Raghu Ganti, Mudhakar Srivatsa\\
IBM Research\\
}
\maketitle

\begin{abstract}
   This technical report describes the design and training of novel speculative decoding draft models, for accelerating the inference speeds of large language models in a production environment. By conditioning draft predictions on both context vectors and sampled tokens, we can train our speculators to efficiently predict high-quality n-grams, which the base model then accepts or rejects. This allows us to effectively predict multiple tokens per inference forward pass, accelerating wall-clock inference speeds of highly optimized base model implementations by a factor of 2-3x. We explore these initial results and describe next steps for further improvements.
\end{abstract}

\section{Introduction}
  As generative Large Language Models (LLMs) grow ever larger, so too do their capabilities. However, this growth in model size in exchange for expressive power presents a problem for deployment in production settings: as powerful as they are, large language models incur substantial memory and computation overhead. Well-known models such as Llama2-13B contain 13 billion parameters, occupying roughly 24 Gb in memory using typical 16-bit weight representations. At inference time, this substantial computational mass is devoted entirely toward producing one single token index at a time, in the range 0 to 32k. Thus in the extreme case, we are incurring a 24 gigabyte overhead to produce less than 2 bytes of information!

  An obvious way to rectify this imbalance would be to predict multiple tokens at a time. Indeed, classical NLP theory has proven that even a simple 2/3-gram language model has great predictive capability, which tells us that learned language models should be capable of predicting more than one token at a time with a reasonable accuracy. However, we'd also like to refrain from altering the base models themselves, as well-known models -- such as (again) Llama-13B -- are known quantities, with desirable properties that cannot be risked in a production setting, where predictability and reliability are paramount. 

  One approach to squaring this circle is \textit{speculative decoding}, where a smaller \textit{draft model} or \textit{speculator} is trained to predict multiple tokens given a sequence of input. These speculative tokens are produced with low cost, and lower accuracy than the base LLM. However, we can leverage GPU parallelism during the LLM forward pass to evaluate the output for each of these new tokens with minimal additional overhead. Then, by comparing the outputs to the speculated inputs, we can accept all the predicted tokens that match the output of the base model, while rejecting all those that don't. In this way we can predict multiple tokens per LLM forward pass at minimal extra cost. A deeper explanation of speculative decoding can be found in \cite{medusa, specu}.

  This paper describes recent efforts at IBM Research to accelerate production LLM inference using speculative decoding with novel architectures. While prior approaches use separate LLMs as the draft model, consuming the entire sequence of token inputs~\cite{specu, tree2}, Medusa~\cite{medusa} showed that a simple multi-headed MLP architecture can reach the same performance. This architecture uses only the base model's embedding vector from the most recent token, as this embedding vector implicitly contains the contextual information derived from the preceding input. We introduce an improved architecture based on Medusa, with the following contributions:

  \begin{enumerate}
      \item We show that speculator output quality can be greatly improved by conditioning on sampled tokens, in addition to the base model context vector.
      \item We introduce an efficient two-stage training scheme, aligning our speculators first to base model input behavior, then to output behavior.
      \item Using this speculator training pipeline, we accelerate four highly optimized production LLMs by a factor of 2-3x.
      \item We explore the limitations of speculative decoding in a production setting, showing that the promised speedups diminish as baseline computation and efficiency levels increase.
      \item We outline next steps and further areas of investigation.
  \end{enumerate}

  Code for our work is open-sourced and available at \url{https://github.com/foundation-model-stack/fms-fsdp}, and the speculators for our 13B-parameter base models are released on HuggingFace: \url{https://huggingface.co/ibm-fms}.

\section{Related Work}
  The original implementations of speculative decoding rely on a smaller, independent LLM to provide drafts for the larger base model~\cite{specu, tree2}. While functional, this approach carries some practical limitations. First, by consuming the entire past token history, even small language models can quickly balloon to large overheads. Second, the smaller models are trained on ground-truth text, and are therefore not aligned to the behavior of the larger base model. Thus the draft LLM could produce suffix candidates that, while reflective of ground truth training data, are inaccurate with respect to the base model behavior, which is the behavior we ultimately want to reproduce. And third, if a smaller LLM trained on the same domain, with the same tokenizer, is not available, training it from scratch might involve unreasonable time and effort.

  The recent Medusa~\cite{medusa} model aims to address all three of these problems by producing drafts via a single, multi-headed MLP. This MLP takes as input the final embedding vector for the base model's most recent token prediction. This allows it to access semantic context without needing the full token history, as the base model has already put in the effort of contextualizing this token vector, making Medusa fast at inference time for any length of prompt. Medusa is also relatively quick and easy to train, as it is a simple model with input that is already semantically rich.

  We identify a major shortcoming in the Medusa architecture: by using only a context vector from the base model, Medusa fails to condition on sampled token selections, leading to poor quality n-gram predictions. EAGLE~\cite{eagle}, Hydra~\cite{hydra} and Recurrent Drafter~\cite{recurdraft} are concurrent works, independently arriving at the same insight. 
  This validates our speculator design, while in contrast to those works, we examine specifically the performance of such models in highly optimized production settings.
  In particular, our base 7B model without acceleration produces 70\% more tokens per second than the equivalent 16-bit precision baseline in the EAGLE case study (94.9 tokens/s vs 55.1~\cite{eagle}), about 90\% higher than Hydra's baseline (roughly 50 tokens/s~\cite{hydra}), and 170\% that of Recurrent Drafter (35.6 tokens/s~\cite{recurdraft}). This leads to a very different - and more interesting - deployment and evaluation landscape.

\section{Speculator Design and Training}
  In this section we describe our speculator architecture, and then discuss efficient training schemes. 

  \subsection{Speculator Design}
  We started with the Medusa architecture, implementing three parallel MLP prediction heads that predict tokens $n+1, n+2, n+3$ using the embedding vector from the base model used to predict token $n$. Looking at the predictions, however, we quickly identified a concerning behavior: the candidate sequences predicted by this Medusa architecture were often very poor trigrams. This is because the Medusa MLP uses the base model embedding vector for token $n$ as its sole input, and completely ignores the actual token sampled from the distribution produced by that vector. This means that the predictions of Medusa for tokens $n+1, n+2, n+3$ are not actually conditioned on token $n$ at all, but rather on the expectation across all possible samples of token $n$, simultaneously. Similarly, token $n+2$ is not conditioned on token $n+1$, but instead on the expectation across all values for $n+1$, and similarly for token $n+3$. This leads to counterintuitive behavior.

  As a simple contrived example, imagine that the base model has produced an input ending with, ``follow these steps:$\backslash$n''. Let the three most likely continuations of this input be (without too much suspension of disbelief required): 
  \begin{itemize}
      \item ``Step 1:''
      \item ``$\backslash$n Step 1:''
      \item ``$\backslash$n $\backslash$n Step 1:''
  \end{itemize}
  Moreover, in each of these continuations, "Step" is highly likely (as we know it must appear), while "$\backslash$n" is likely to have lower confidence. In this scenario, the Medusa prediction for the next three tokens is then ``Step Step Step'', as those are clearly the three most likely tokens at each position $n+1,n+2,n+3$. However, this is also clearly a poor predictor of base model behavior.

\begin{figure}[t]
       \centering
       \includegraphics[width=0.5\linewidth]{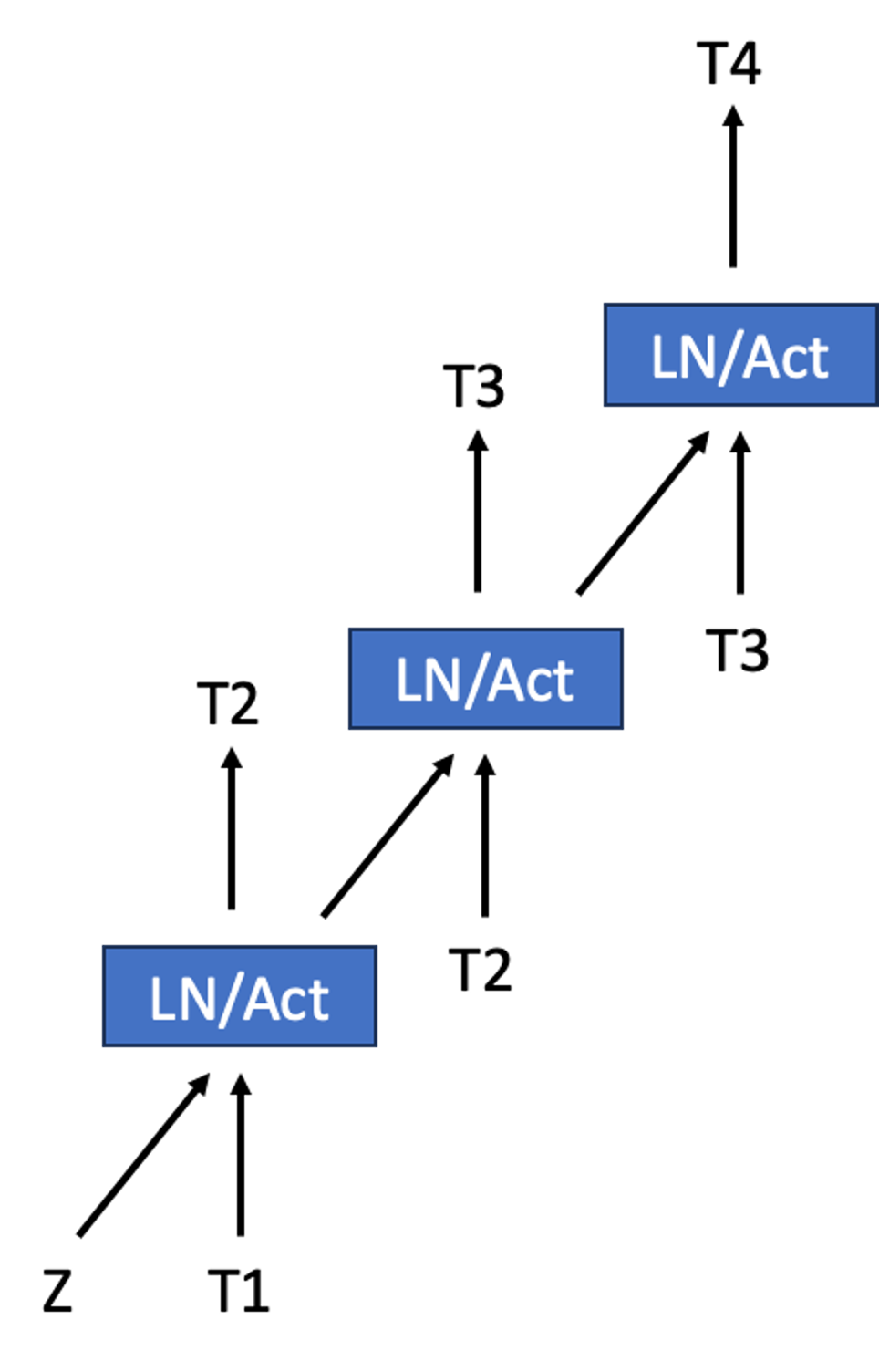}
       \caption{A simple architecture diagram for a 3-headed multi-stage MLP speculator. Z is the latest state vector from the base model, while T[n] is the sampled token at time $t+n$.}
       \label{fig:arch}
    \end{figure}

  To alleviate this problem, we modify the Medusa architecture into a multi-stage MLP, where each stage predicts a single token in the draft sequence. Each stage of the MLP takes as input both a state vector and one (or more) embeddings for the sampled token(s) from the prior stage (the base model can be considered stage 0). Each input pair (state plus sampled token embedding) is projected and passed through a LayerNorm/GeLU activation, forming a new state vector. This state vector is used to predict the next set of draft tokens, which, with the new state vector, acts as the inputs for the next stage of prediction. The model thus emits an entire tree of candidates, and we allow the user to select the top-k most likely predictions from that tree to use in practice. The 3-stage model architecture is visualized in Fig.~\ref{fig:arch}.

  At inference time, we implement multi-candidate decoding~\cite{multicandidate} and Tree Attention~\cite{medusa, tree2} to evaluate the chosen $k$ candidates simultaneously in parallel. Rather than compare input tokens to output tokens for a single candidate sequence, and taking only the correct tokens, we instead perform this scoring process for all candidates simultaneously, and keep the tokens from only the best-scoring candidate. This is a simple way to increase speculator accuracy (and thus the number of generated tokens per step), at the cost of increased parallelism. When we report tokens per step results, therefore, we will also report the number of candidates being evaluated in parallel. 

  By conditioning on prior tokens as well as the base model's internal state, our speculator is able to produce more realistic n-gram drafts. At the same time, it's also more accurate at predicting base model behavior. For Llama2-13B, we were able to accelerate the above three-headed sandbox Medusa architecture from 2.28 tokens per step, to 2.63 tokens per step (best of 10 candidates), simply by conditioning each MLP head on the previous sampled tokens, in addition to the base model state.

  Note that Fig.~\ref{fig:arch} could just as easily depict an RNN, and indeed we could implement our multi-stage MLP as an RNN by simply tying the weights across the different stages. While this weight sharing would shrink the parameter count of our speculator, it does \textit{not} make the speculator any faster at inference time. Therefore we implement our speculator without any weight sharing for maximum adaptability. The resulting parameter counts thus appear large (800M-1.1B for a 7B-13B scale base LLM). But despite having $\sim1/10$th the parameters of the base LLM, our speculators execute in under $1/30$th of the time, due to their shallow depth and simple architecture.

  \subsection{Speculator Training}

  To train our speculators, we leverage the efficient Foundation Model Stack distributed training platform introduced in~\cite{fsdpblog}. We adapt the Llama training script to instead freeze the base Llama model and instead train the speculator on top. We keep  the same dataloaders, checkpoint utilities, and other convenient pipeline components.

  Since the speculator is trained from scratch, we have the opportunity to fully align it to the base model: instead of predicting ground truth text, the speculator should train directly against the generated base model output. Unfortunately, text generation with Llama is not efficiently parallelizable -- the whole reason we're training a speculator in the first place. We therefore adopt a 2-stage training paradigm. In the first stage, we train on small batches with long context lengths (4k tokens). The base model generates embedding vectors for each token in parallel, and the speculator uses these vectors, plus the ground truth tokens, to predict subsequent ground truth tokens. This is a similar task to standard causal language model training, only we're predicting multiple tokens at each position (one for each stage of the speculator MLP).

  In stage 2, we switch to large batches with short sequence lengths (256 tokens), generated from the base model itself. By only generating a short sequence of tokens, we can continue to train reasonably efficiently in parallel, while fine-tuning the speculator to explicitly match base model outputs. We did not observe any consistent overfitting to short sequence lengths -- in fact, in some cases, the speculator became more accurate on longer inputs. In our setup, a roughly 5:2 ratio of steps for stage 1 vs stage 2 translates to a roughly 1:1 ratio in wall clock time, so we use this ratio for convenience. The speculator efficiently trains on proxy ground-truth text data for half the time, and then spends the other half on less efficient model alignment.

\begin{figure}[t]
       \centering
       \includegraphics[width=\linewidth]{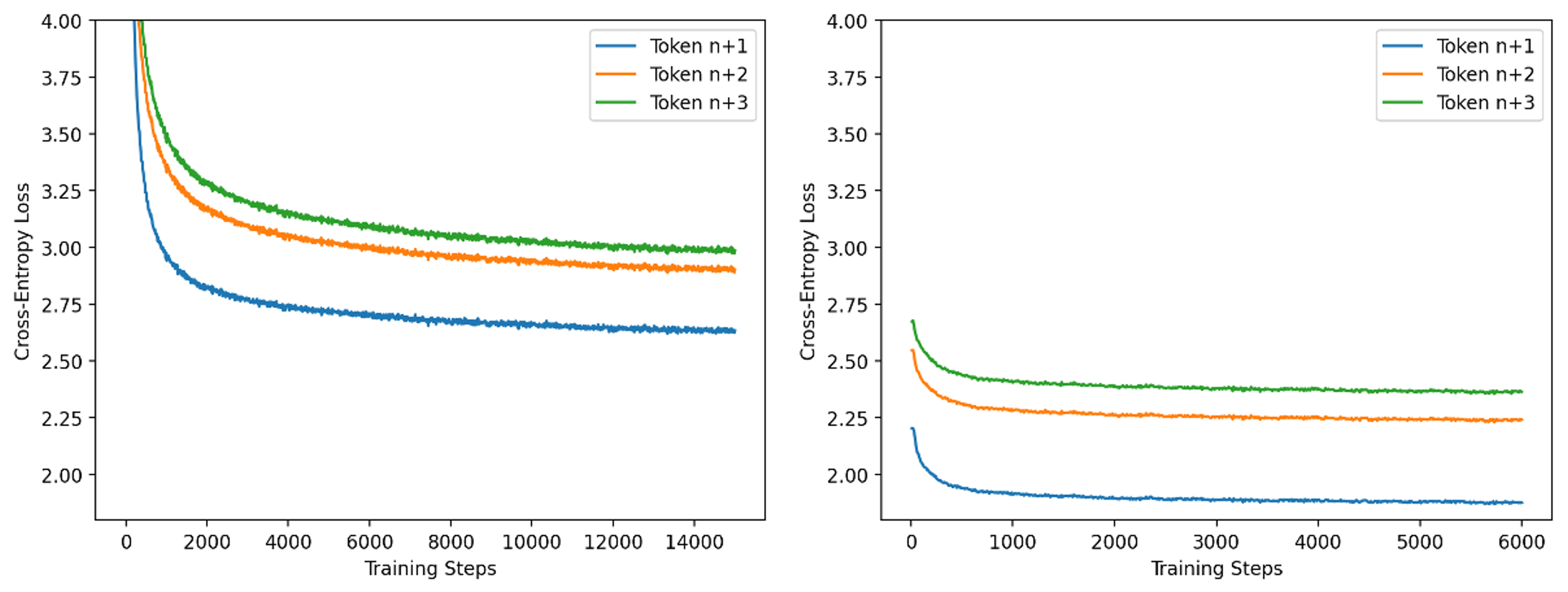}
       \caption{Per-head training loss curves for Llama2-13B speculator training, stages 1 (left) and 2 (right). Loss values jump downward at the start of stage 2 as the task becomes inherently easier: rather than using base model behavior to predict ground truth text, we are using base model behavior to predict other base model behavior (future tokens).}
       \label{fig:loss}
    \end{figure}

  By the end of stage 2 training, we are able to accelerate a three-headed Llama2-13B speculator from 2.41 tokens per step at the end of stage 1, to 2.63 (best of 10 candidates). The training curves for this run across both training stages are provided in Fig.~\ref{fig:loss}.

\section{Results}
  We train four rounds of speculator models, and discuss the results for each. The first is a proof-of-concept 3-headed speculator for Llama2-7B, which we use to illustrate both the successes and limitations of our approach in production settings. The second is a similar 3-headed speculator for Llama2-13B, which we deploy to an internal LLM server. The third is a 7-headed speculator for Codellama-13B-instruct, which achieves much better speedup due to the more predictable nature of code. Finally, we train a 5-headed speculator for an internal Granite-20B code model, demonstrating further scalability with model size.
  
  All speculators are trained on 32 NVIDIA A100 gpus, with latent dimensionality 4096, 15k steps for stage 1 training, and 6k for stage 2. For natural language models, stage 1 batches contain 1M tokens, and stage 2 batches consist of 786M tokens generated from 197M tokens' worth of truncated training documents. Because the code models are more computationally involved (7 or 5 heads vs 3) we shrink batch sizes to 786M tokens for stage 1, and 524M/131M generated/ground truth tokens for stage 2. All layers are initialized to standard deviation $4096^{-1/2}$, and state/embedding projections are weighted such that the base model's state vector input accounts for $50\%$ of the variance of the state at the final head, in expectation. Learning rate is set to $1e-3$ for stage 1, with warmup over the first $5\%$ of training and cosine decay to $1e-4$. Stage 2 follows the same LR scheme, but with a max LR of $1e-4$ and annealing to $1e-5$, as we found that higher values caused loss to spike badly during the transition from stage 1 to 2.

  \subsection{Llama2-7B-chat}
  In Table~\ref{tab:7b} we track the per-token latency of greedy inference with a Llama2-7B model and our trained speculator on a single A100, in fp16 precision, as a function of batch size ($b$), prompt length ($p$), and number of simultaneous candidates ($k$). The speculator generates a tree with 6 branches at the first head, 3 at the second, and 2 at the third, and we prune this tree of $6*3*2=36$ candidates down to the desired $k$ base on confidence. For reference, we include non-speculative inference under the heading $k=0$, and the average tokens generated per step ($\tau$) across all batch sizes. Latencies are calculated over 256 trials of exactly 100 generated tokens. Prompts are Webhose documents truncated to the desired prompt length.

    \begin{table*}
      \centering
      \scriptsize
      \begin{tabular}{c|ccccccc}
        \toprule
        ms/tok (64) & 0 & 1 & 2 & 4 & 8 & 16 & 32 \\
        \midrule
        1 & 10.54 & 6.23 & 5.76 & \textbf{5.50} & 5.55 & 6.00 & 7.21 \\
        2 & 10.66 & 6.32 & 6.01 & 5.91 & 6.32 & 7.62 & 11.15 \\
        4 & 10.78 & 6.78 & 6.67 & 6.95 & 8.73 & 11.90 & 18.31\\
        \midrule
        $\tau$ & 1.00 & 2.21 & 2.44 & 2.67 & 2.84 & 2.96 & 3.03\\
        \bottomrule
      \end{tabular}
      \hspace{.3in}
      \begin{tabular}{c|ccccccc}
        \toprule
        ms/tok (2048) & 0 & 1 & 2 & 4 & 8 & 16 & 32 \\
        \midrule
        1 & 12.87 & 9.11 & \textbf{8.89} & 8.91 & 9.68 & 11.51 & 14.78 \\
        2 & 14.98 & 10.78 & 10.94 & 11.56 & 13.24 & 16.83 & 24.96 \\
        4 & 19.62 & 14.24 & 15.63 & 16.78 & 20.97 & 29.05 & 44.88 \\
        \midrule
        $\tau$ & 1.00 & 2.01 & 2.21 & 2.42 & 2.61 & 2.74 & 2.83 \\
        \bottomrule
      \end{tabular}
      \caption{Iterative latency (milliseconds per token) for Llama2-7B. Rows indicate batch size $b$, columns indicate number of parallel candidates $k$. $k=0$ indicates non-speculative baseline, and logical speedup (tokens per step) is given as $\tau$. Prompt length $p$ is 64 (left) vs 2048 (right). As baseline computational load increases, speculative decoding provides less improvement.}
      \label{tab:7b}
    \end{table*}

  From the table we can make a few notable observations. First, our logical speedups ($\tau$, increasing asymptotically with $k$) do translate to wall-clock speedups, but not one to one. This makes sense in a production setting, as our inference implementation is already highly optimized and torch compiled~\cite{compileblog}. Our baseline $b=1,k=0$ latency of 10.54 milliseconds per token translates to 94.9 tokens per second for Llama2-7B, compared to EAGLE's gpt-fast implementation baseline of 55.1 tokens per second~\cite{eagle}. Note that speculative decoding relies on unutilized GPU bandwidth to parallelize the evaluation of multiple candidates and tokens during the base model forward pass. If GPU throughput is already maximized, that leaves less resources to dedicate toward speculative decoding based acceleration. Here we observe just under 2x speedup for prompt length 64, batch size 1, and 4 candidates per step. This is less than our measured tokens per step (2.67), but still a massive improvement in a production setting!

  Second, and consistent with the above intuition on limited GPU bandwidth: as prompt length and batch size increase, the speedup from speculative decoding erodes. For large prompts and batch sizes, speculative decoding can be over \textit{twice} as slow as non-speculative decoding, despite generating almost three tokens per single forward pass. This is less surprising when we consider that for $b=4,k=32$ and 4 tokens per candidate (1 ground truth token from the base model plus 3 speculated suffix tokens), we're running 512 tokens per step. In terms of raw FLOPs, the equivalent non-speculative workload is a batch size of 512, far beyond the capabilities of an A100 gpu, which we observed failing at $b\approx128$ due to memory limits. Thus while a promising technique for inference acceleration, the wall-clock speedup from speculative decoding scales inversely to both the computational workload and the degree of pre-existing optimization of the base model.

  Finally, because of this falloff in performance, the optimal number of parallel candidates for a given batch size is usually quite small in our setup: less than 8 in all cases for prompt length 64, and less than 4 when prompt length is 2048. This is in contrast to prior work in speculative decoding, which usually features larger sets of candidates. For example, Lookahead Decoding~\cite{lookahead} obtains their best results for $k=15$, while \cite{multicandidate} publishes results for $k=32$. In our case, inflating the input tensor to this size hurts more than helps, and also calls into question the contribution of tree attention in our implementation, as there are likely to be very few redundant tokens when $k$ is only 4 or less.

  In sum, the Llama2-7B speculator results indicate that speculative decoding produces significant material speedups in production settings. However, the practitioner must carefully account for computational load. The choice of hyperparameters and even algorithms may require dynamic adaptation to workloads at runtime.

  \subsection{Llama2-13B-chat}
\begin{figure}[t]
       \centering
       \includegraphics[width=.8\linewidth]{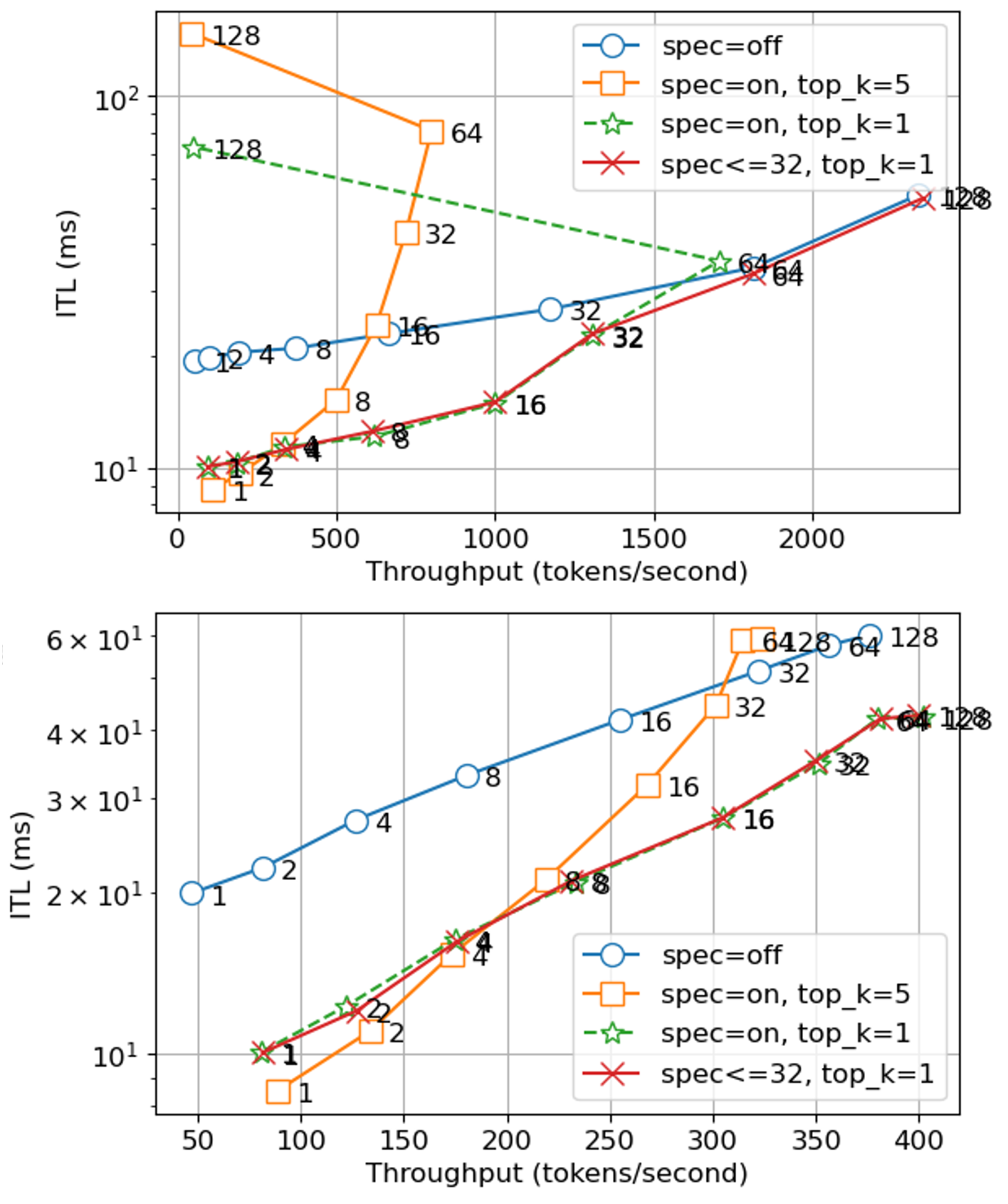}
       \caption{Throughput (x-axis) vs iterative token latency (y-axis) for Llama2-13B with speculative decoding on an inference server. Numbers indicate concurrent users. Top: artificial homogeneous workloads (49 tokens in, 100 tokens out). Bottom: heterogeneous workloads matched to historical logs. Speculators scale better in the heterogeneous case (but at lower throughput) because batch size is upper-bounded by concurrent users $b$, but in practice is lower in expectation.}
       \label{fig:tgis}
    \end{figure}

  Moving beyond proof of concept in Llama2-7B, we train a new speculator, using the same hyperparameters, for Llama2-13B. Speculator architecture remains the same, except that the initial state vector input now has dimensionality $5120$ instead of $4096$, as the base model width has increased. We then deployed the accelerated base model to an internal IBM model inference server. Benchmarks of latency and throughput for $k=[1,5]$ and varying $b$ are shown in Fig.~\ref{fig:tgis}. We observe a similar 2x reduction in latency for $b=1$ and $k=5$. However, consistent with the observed behavior for 7B, this improvement gradually disappears for larger $b$, until speculative decoding with $k=5$ becomes significantly slower than the baseline. 

  Fortunately, we can mitigate this behavior by fixing $k=1$, trading off a slightly higher latency for $b<4$ for much better behavior at $b>4$. In the extreme case, we can also disable speculative decoding dynamically for $b$ above a given threshold, such as 32. In this way, we can attain $>2x$ speedups in the best case, and perfectly reproduce baseline behavior in the worst case.
  These results suggest that in large-scale production contexts, a well-trained speculator may be most performant with $k=1$, and no special multi-candidate handling, as the marginal improvement in recall is not worth incurring the overhead of evaluating extra candidates.

  \subsection{Codellama-13B-instruct}
  Shifting to a different data domain, we train a speculator for Codellama-13B-instruct. Because code is more formalized, repetitive, and therefore predictable than natural language, we increase the number of stages/heads in our speculator from 3 to 7 (raising input length from 4 to 8) to capitalize on this predictability. Performance benchmarks are given in Table~\ref{tab:code} for prompt lengths 64 and 512. 

    \begin{table*}
      \centering
      \scriptsize
      \begin{tabular}{c|cccccc}
        \toprule
        ms/tok (64) & 0 & 1 & 2 & 4 & 8 & 16 \\
        \midrule
        1 & 18.88 & 6.3 & 5.99 & 5.67 & 5.76 & 6.96 \\
        2 & 19.34 & 9.01 & \textbf{5.51} & 5.64 & 7.22 & 8.71 \\
        4 & 19.60 & 5.55 & 5.94 & 7.86 & 10.08 & 13.99 \\
        \midrule
        $\tau$ & 1.00 & 5.02 & 5.31 & 5.56 & 5.76 & 5.86 \\
        \bottomrule
      \end{tabular}
      \hspace{.3in}
      \begin{tabular}{c|cccccc}
        \toprule
        ms/tok (512) & 0 & 1 & 2 & 4 & 8 & 16 \\
        \midrule
        1 & 20.12 & 6.46 & 6.44 & 10.55 & 6.60 & 7.98 \\
        2 & 21.14 & 6.25 & \textbf{6.19} & 6.90 & 9.03 & 10.93 \\
        4 & 22.82 & 6.71 & 7.55 & 10.16 & 13.43 & 17.91 \\
        \midrule
        $\tau$ & 1.00 & 5.49 & 5.73 & 5.92 & 6.10 & 6.21 \\
        \bottomrule
      \end{tabular}
      \caption{Iterative latency (milliseconds per token) for Codellama-13B-instruct. Rows indicate batch size, columns indicate number of parallel candidates. Candidates 0 indicates non-speculative baseline, and logical speedup (tokens per step) is given as $\tau$. Prompt length is 64 (left) vs 512 (right). As baseline computational load increases, speculative decoding provides less improvement.}
      \label{tab:code}
    \end{table*}

  Here we observe that the increased heads are suitable for code workloads. By doubling the maximum acceptable n-gram length, we roughly double the mean accepted n-gram length $\tau$. This $>5x$ logical speedup produces a $>3x$ wall-clock speedup, allowing us to operate a 13B parameter model at fp16 precision with a throughput of \textit{181.5 tokens per second} ($5.51$ ms/token at $b=2,k=2,p=64$). 

  Beyond the notably improved performance, results in Table~\ref{tab:code} largely mirror those in Table~\ref{tab:7b}. Improvement falls off as batch size and prompt length increase, shrinking the optimal $k$, which remains small in general. A discrepancy is the fact that speculator accuracy, as indicated by $\tau$, actually increases slightly with prompt length, as opposed to decreasing slightly in Table~\ref{tab:7b}.

  \subsection{Granite-20B}
\begin{figure}[t]
       \centering
       \includegraphics[width=.8\linewidth]{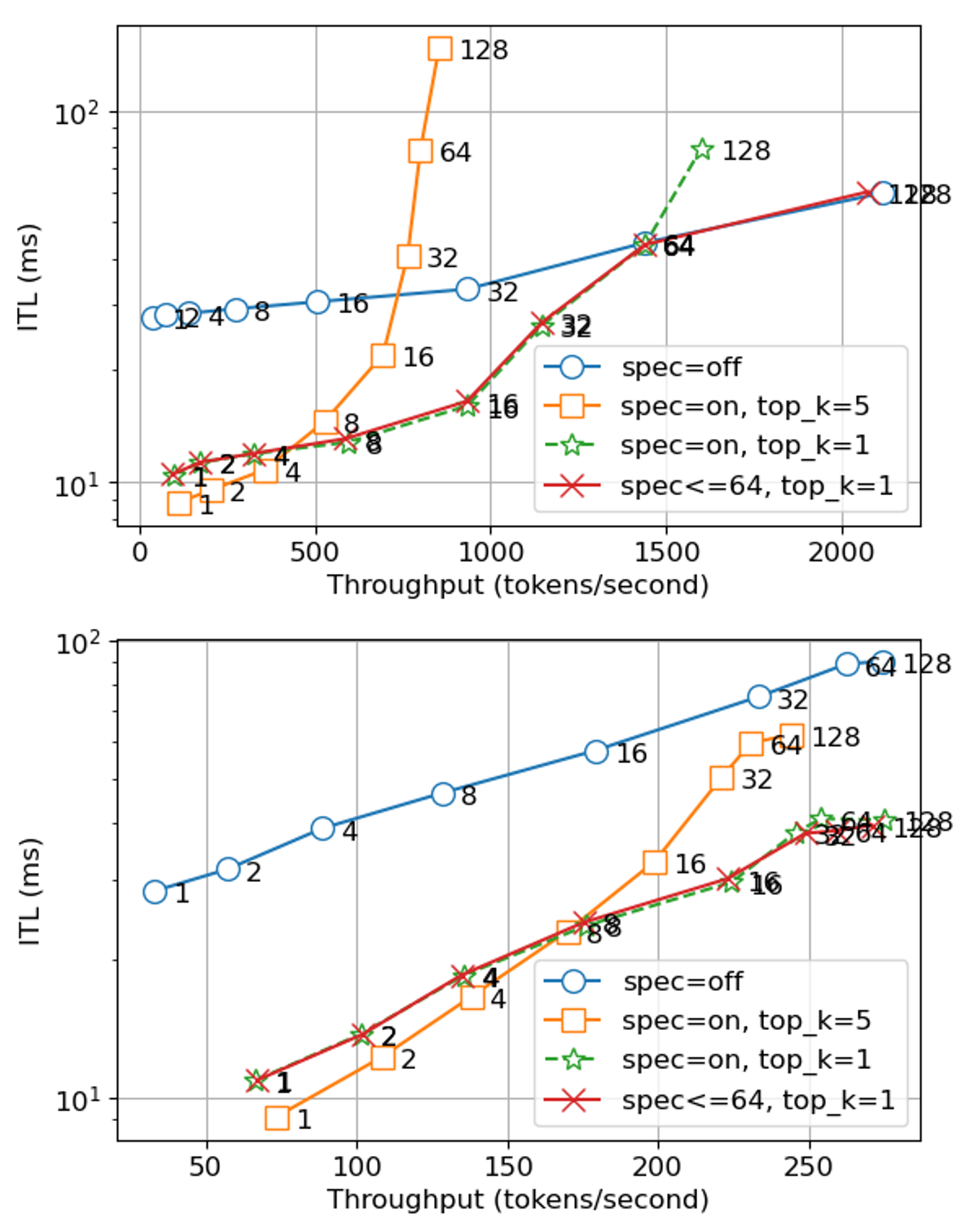}
       \caption{Throughput (x-axis) vs iterative token latency (y-axis) for Granite-20B with speculative decoding on an inference server. Numbers indicate concurrent users. Top: artificial homogeneous workloads (50 tokens in, 100 tokens out). Bottom: heterogeneous workloads matched to historical logs. Speculators scale better in the heterogeneous case (but at lower throughput) because batch size is upper-bounded by concurrent users $b$, but in practice is lower in expectation.}
       \label{fig:granite}
    \end{figure}

  Finally, we train a speculator for a larger, internal code model, Granite-20B. Due to memory constraints at training time, we reduce the number of speculator heads in our code speculator from 7 to 5, and otherwise train the same as for Codellama-13B. Similar to our Llama2-13B speculator, we then deploy this model to our internal server and benchmark latency and throughput across batch sizes, with results given in Fig.~\ref{fig:granite}. 

    Behavior mirrors prior results, despite the larger model size. Consistent with Codellama-13B, we see a roughly $3x$ speedup over the non-speculative baseline in the best case. Similar to Llama2-13B, we find that fixing $k=1$ buys us much better scaling behavior, at the cost of a slight latency drop for very small batch sizes. Heterogeneous workloads scale better than homogeneous ones, though with lower throughput, due to smaller batch sizes in practice (simulated users no longer artificially synchronize their calls). And by disabling speculative decoding for large batch sizes, we can achieve the best of both worlds.

\section{Discussion}
  While achieving a 2-3x speedup in a production setting is significant, the approach discussed so far represents a version 1 first attempt. A number of avenues are available to explore for further improving the efficiency of our speculator models.

  First, our natural language speculators are constrained by their limited number of heads. Because we only speculate 3 tokens at a time, we are capped at 4 tokens per step (1 ground truth prediction from the base model, plus up to 3 accepted speculated tokens). Increasing the number of heads to 4, as in \cite{hydra}, or 5, as in \cite{eagle} and \cite{recurdraft}, is a simple and easy way to improve the performance of our models. The same argument may also be applied to our 5-headed Granite-20B code model, though the fact we attain the same wall-clock latency improvement as for our 7-headed Codellama-13B speculator, suggests that heads 6 and 7 may be redundant. Further study is needed.

  Second, we declined to tie any weights in our speculators, either between heads/stages, or between the speculator and the base model. This is because doing so provides no wall-clock speedup. However, decreasing the parameter count via weight tying can be useful in terms of shrinking GPU memory overhead, especially as we increase the number of heads/stages in our speculators. It might also lead to improved convergence, for example by initializing the embedding layers or shared layer to the semantically rich values of the base model's embedding layer. Working in the same latent space as the base model also opens opportunities for introducing useful auxiliary losses, as in \cite{eagle}. 

  At inference time, the optimal $k$ depends on $b$, and so we'd like to swap dynamically to the optimal value for each incoming inference request, step to step, rather than simply toggling on/off between default values of $k=(0,1,5)$. However, the optimal scaling plan will be model- and hardware-specific, and thus, though desirable for an efficient speculative decoding implementation, unlikely to yield many broader insights. 

\section{Conclusion}
  In this technical report we introduced a novel improvement on the Medusa speculative decoding architecture, improving predicted token sequences by conditioning on prior sampled tokens, in addition to the base model's current state vector. Using a two-stage training process, we produce trained accelerators for Llama2-7B, Llama2-13B, Codellama-13B, and an internal Granite-20B model, and discuss performance impacts on a highly optimized production-level baseline. In particular, speculative decoding delivers its promised speedups in optimal scenarios, and we are able to run blazingly fast at fp16 precision. However, the improvement degrades for larger workloads, as the available window for handling increased parallelism on the GPU shrinks. Thus, delivering reliable acceleration via speculative decoding in production settings requires thoughtful -- and careful -- design and effort. 

\section{Acknowledgments}
We thank Supriyo Chakraborty, for insightful design discussions in the early stages of the project, and Antoni Viros i Martin, Linsong Chu, and Brian Vaughan, for valuable consultations on code.

\newpage
\newpage

{\small
\bibliographystyle{ieee_fullname}
\bibliography{egbib}
}

\end{document}